\documentclass{article}

\usepackage{arxiv}
\usepackage{amsmath,amssymb,amsfonts}
\usepackage[utf8]{inputenc} 
\usepackage[T1]{fontenc}    
\usepackage{hyperref}       
\usepackage{url}            
\usepackage{booktabs}       
\usepackage{amsfonts}       
\usepackage{nicefrac}       
\usepackage{microtype}      
\usepackage{lipsum}		
\usepackage{graphicx}
\usepackage{natbib}
\usepackage{doi}

\title{KemenkeuGPT: Leveraging a Large Language Model on Indonesia's Government Financial Data and Regulations to Enhance Decision Making}

\author{ 
{\hspace{1mm}Gilang Fajar Febrian}\\
	School of Computer Science\\
	University of Nottinghsm\\
	Nottingham NG7 2RD, UK\\
	\And
	{\hspace{1mm}Grazziela Figueredo} \\
	School of Computer Science\\
	University of Nottinghsm\\
	Nottingham NG7 2RD, UK\\
}

\date{}

\renewcommand{\headeright}{}
\renewcommand{\undertitle}{}
\renewcommand{\shorttitle}{}

\hypersetup{
pdftitle={KemenkeuGPT: Leveraging a Large Language Model on Indonesia's Government Financial Data and Regulations to Enhance Decision Making},
pdfsubject={Computer Science, Artificial  Intelligence},
pdfauthor={Gilang Fajar Febrian, Grazziela Figueredo},
pdfkeywords={Large Language Models (LLMs), LangChain, Retrieval-Augmented Generation (RAG), prompt engineering, fine-tuning, Indonesia Government financial data, financial regulations, human feedback.},
}

\begin{document}
\maketitle

\begin{abstract}
Data is crucial for evidence-based policymaking and enhancing public services, including those at the Ministry of Finance of the Republic of Indonesia. However, the complexity and dynamic nature of governmental financial data and regulations can hinder decision-making. This study investigates the potential of Large Language Models (LLMs) to address these challenges, focusing on Indonesia's financial data and regulations. While LLMs are effective in the financial sector, their use in the public sector in Indonesia is unexplored. This study undertakes an iterative process to develop KemenkeuGPT using the LangChain with Retrieval-Augmented Generation (RAG), prompt engineering and fine-tuning. The dataset from 2003 to 2023 was collected from the Ministry of Finance, Statistics Indonesia and the International Monetary Fund (IMF). Surveys and interviews with Ministry officials informed, enhanced and fine-tuned the model. We evaluated the model using human feedback, LLM-based evaluation and benchmarking. The model's accuracy improved from 35\% to 61\%, with correctness increasing from 48\% to 64\%. The Retrieval-Augmented Generation Assessment (RAGAS) framework showed that KemenkeuGPT achieved 44\% correctness with 73\% faithfulness, 40\% precision and 60\% recall, outperforming several other base models. An interview with an expert from the Ministry of Finance indicated that KemenkeuGPT has the potential to become an essential tool for decision-making. These results are expected to improve with continuous human feedback.
\end{abstract}


\section{Introduction}
Data is essential for comprehending issues, engaging the public, and enhancing public services. Furthermore, it fosters an environment that promotes robust, evidence-based decision-making [\citealt{van2019data}]. The Minister of Finance of the Republic of Indonesia, Sri Mulyani Indrawati, highlighted that the Ministry of Finance possesses vast data, which she described as a new type of data mine for the digital era [\citealt{canrakerta2021membangun}]. However, the vast and varied nature of governmental financial data and regulations makes manual collection and analysis difficult. Frequent changes in government rules and continuous updates to financial data further hinder access to real-time information, thereby delaying decision-making. Currently, the ministry depends on a dashboard that offers financial data information and a finance regulation information network website to find information about regulations. However, gathering new data not already available on the dashboard and manually searching for regulations is time-consuming. Therefore, there is a need for artificial intelligence (AI) systems that can extract insights from Indonesian financial data and regulations automatically. To address these challenges, LLMs could become a solution for analysing and extracting information from Indonesian government financial data and regulations. The state-of-the-art literature of LLMs in the financial domain mostly employs fine-tuning in financial private sectors, which are trained with corporation financial data. However, this approach, which relies solely on fine-tuning, can generate misleading responses or hallucinations. This drives the motivation for this study to propose KemenkeuGPT, a continuous improvement LLM application using LangChain through RAG, prompt engineering and fine-tuning.

The main objectives of this research are to:
\begin{itemize}
    \item Develop KemenkeuGPT, an LLM application with LangChain and RAG.
    \item Assess the performance of several pre-trained models on Indonesian financial data and regulations.
    \item Engage with stakeholders in the Ministry of Finance through surveys and interviews to obtain feedback on KemenkeuGPT’s performance.
    \item Improve the performance of KemenkeuGPT by adding new data based on stakeholder feedback, prompt engineering, and fine-tuning.
    \item Evaluate KemenkeuGPT's performance by measuring its accuracy, correctness, faithfulness, precision, and recall.
    \item Assess the performance of KemenkeuGPT in aiding decision-making by interviewing experts in the Ministry of Finance.
    \item Implement a multi-platform interface for KemenkeuGPT application to enhance user experience.
\end{itemize}

This paper is divided as follows. In the second section, we survey existing research covering topics including government financial data and regulations in Indonesia, data-driven decision-making in finance and LLMs in finance. The third section details system development of LangChain framework with RAG approaches. We also discuss RAG, prompt engineering and fine-tuning. Subsequently, we provide an overview of the research design conducted in this paper. The fourth section provides a comprehensive overview of the iterative improvement and development process, including experiments with LLMs, RAG, prompt engineering, fine-tuning and human feedback. Furthermore, we explain the evaluation methods, including human evaluation, LLM-based and benchmarking. In section five, we present the results and discussion of the experiments. Section six concludes and summarises the research process, including results, limitations and future work.

\section{Background}

\subsection{Related work}

Several instances of chatbot implementation exist in the public finance sector. The Australian government’s Tax Service chatbot, Alex, and Botty Bonn, a chatbot from a German city government, allow citizens to handle bill payments and taxes [\citealt{chen2024adoption}]. Additionally, a chatbot could be used to explore open government data (OGD), including government financial data [\citealt{porreca2018accessing}]. These chatbots enable users, including non-experts, to interactively search and explore government data through natural language, enhancing the ease of data access and interpretation. This approach significantly improves public service delivery, allowing for more informed decision-making and greater transparency. The benefits of improved accessibility and user-friendliness drive the integration of AI in government operations. With their advanced natural language processing abilities, chatbots offer a more intuitive way for citizens and officials to interact with government data. They are instrumental in managing large volumes of queries, offering real-time responses, and handling complex data explorations. Thus, AI contributes to a more efficient decision-making process.

\subsection{Data Driven Decision Making in Finance}

Data-driven decision-making refers to the practice of making decisions based on data analysis rather than solely relying on intuition [\citealt{provost2013data}]. For example, a government tax officer could set a tax target for the next year based purely on his/her experience and intuition about what might work. Alternatively, he/she could base his/her calculation on data analysis regarding the previous tax target, achievements, and other factors that could influence the tax target. There are four types of analytics that could be employed, namely descriptive, diagnostic, predictive, and prescriptive [\citealt{sarker2021data}]. Descriptive analytics focuses on summarising past data to offer insights into historical trends. Diagnostic analytics goes further by examining the underlying causes of specific outcomes, aiding in root cause analysis. Predictive analytics utilises statistical models to forecast future trends based on historical data, enabling proactive decision-making. Lastly, prescriptive analytics goes beyond predictions, offering recommendations and optimisation strategies to achieve desired outcomes. Data analytics is a crucial concept encompassing the exploration, understanding, and communication of new insights and significant patterns derived from large datasets across various application domains. It aims to facilitate swift, high-quality and efficient decision-making processes [\citealt{ojokoh2020big}]. Despite being in its early stages of development, data analytics is becoming increasingly essential in supporting evidence-based decision-making within the global public sector [\citealt{charles2022artificial}].

\subsection{LLMs in Financial Domain}

LLMs are transformer-based models with hundreds of billions of parameters, trained on extensive datasets, enabling them to understand natural language and address complex tasks effectively [\citealt{zhao2023survey}], such as GPT-4 [\citealt{openai2023gpt}], LLaMA [\citealt{touvron2023llama}], PaLM [\citealt{chowdhery2023palm}], LaMDA [\citealt{thoppilan2022lamda}], Galactica [\citealt{taylor2022galactica}], and BERT [\citealt{devlin2018bert}].Additionally, there are domain-specific LLMs, which are increasingly recognised as crucial for effectively processing textual documents to extract relevant information in specific contexts. This methodological approach has proven to significantly boost the performance of language models, ensuring their accuracy and effectiveness in specific contexts [\citealt{chakrabarty2019imho}]. Among the various applications of domain-specific LLMs, one notable area is the financial domain. Recent work in the financial domain has introduced tools such as FinGPT [\citealt{liu2023fingpt}]. FinGPT is a large language model specifically designed for the financial sector. It aims to democratise financial data and language models by providing an open, accessible framework for researchers and practitioners. This model adopts a data-centric approach, emphasising the importance of data acquisition, cleaning, and preprocessing to develop robust financial applications. This approach demonstrates how LLMs can assist in various financial tasks, from automated advising to trading analysis using news, social media, company filings, and research datasets. Another model, BloombergGPT [\citealt{wu2023bloomberggpt}], has been trained on extensive financial data, including news, filings, press releases, web-scraped financial documents, and social media sourced from Bloomberg archives. BloombergGPT is a specialised large language model with 50 billion parameters designed for the financial sector. It is trained on a vast, domain-specific dataset of 363 billion tokens derived from Bloomberg’s financial data sources. This dataset is supplemented by 345 billion tokens from general-purpose datasets, making it one of the most enormous domain-specific datasets utilised for a language model. Furthermore, FinBERT [\citealt{huang2023finbert}] is a financial sentiment analysis model based on BERT architecture, which has been specially adapted for the financial domain. It utilises BERT’s powerful pre-training and fine-tuning capabilities to specifically tackle natural language processing (NLP) tasks within the financial sector, such as sentiment analysis of financial texts. It has been trained on a vast corpus of financial texts, including corporate filings, analyst reports, earnings conference call transcripts, and CSR reports.

\section{System Development}
This study’s design explores an iterative process for developing an LLM application with LangChain. The research process is designed to be cyclic rather than linear, emphasising continuous improvements through repeated cycles of data collection, processing, development, experiment, evaluation, and refinement, as shown in Figure 1.

\begin{figure*}
	\centering 
	\includegraphics[width=0.75\textwidth]{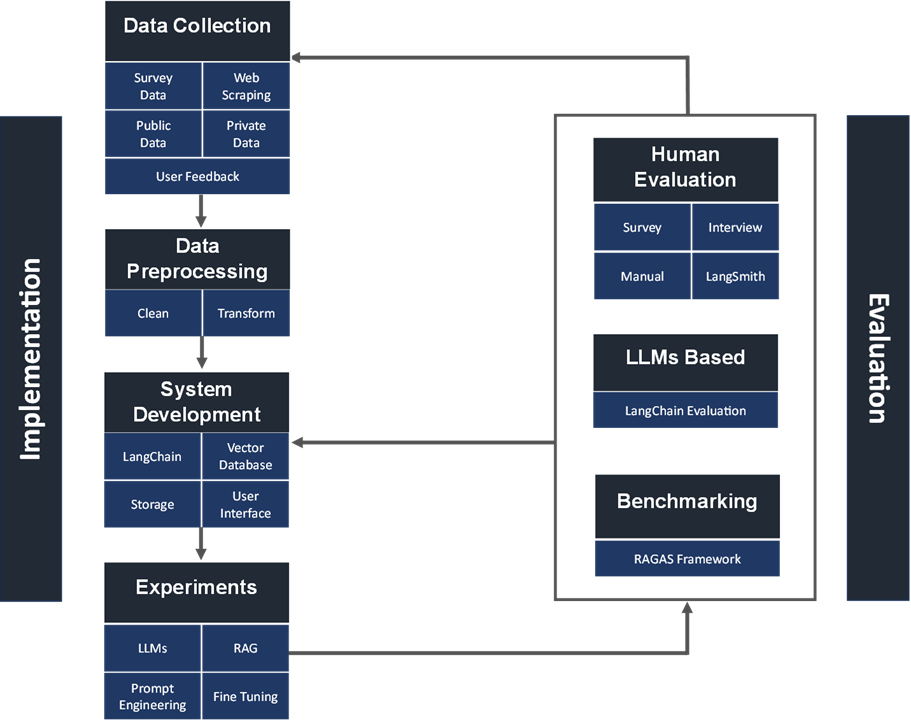}	
	\caption{Iterative development of the research.}
        \label{fig:img1}
\end{figure*}

KemenkeuGPT has been developed with LangChain framework and iterative development process with RAG, prompt engineering, and fine-tuning. LangChain is a framework that streamlines the development of applications that LLMs. As an integration framework for these models, LangChain is applicable in various areas commonly associated with language models, such as analysing and summarising documents, developing chatbots, and examining code. LangChain has shown considerable promise within the AI ecosystem by introducing a novel approach that allows developers to interact seamlessly with various data sources and applications. LangChain’s modular abstractions and customisable, use-case-specific pipelines mark its flexibility and efficiency. These features position it as an invaluable tool for the future development of LLM applications [\citealt{topsakal2023creating}].
In the financial domain, LangChain’s capabilities have been effectively demonstrated in tasks related to financial analysis. A Financial Large Language Model (FLLM) processes the initial corpus to create a link between the analysed input and various knowledge sources, such as expert domain knowledge, financial databases, and search engines. This FLLM, designed for multitask prompt-based fine-tuning, addresses three essential components of financial analysis and interpretation: event matching and analogy, evaluation of viewpoint quality, and extraction of key points. The approach focused on data through LangChain offers a promising direction for leveraging the potential of LLMs in the complex financial domain.
RAG is a method that merges context retrieval capabilities with LLMs for generating language [\citealt{cai2022recent}][\citealt{lewis2020retrieval}]. It functions in a two-phase manner. Initially, it uses a retrieval module to find pertinent documents based on the given prompt, often drawing from external sources such as news outlets, research papers, and social media for added context. Subsequently, these documents are integrated with the initial prompt and input into the LLMs, culminating in the production of the final output.
There is a retrieval-augmented LLM framework for financial sentiment analysis. This framework includes an instruction-tuned LLM module, which ensures the role of LLMs as predictors of sentiment labels and a retrieval-augmentation module that retrieves additional context from reliable external sources. Benchmarked against traditional models and LLMs like ChatGPT and LLaMA, this approach achieves a 15\% to 48\% performance gain in accuracy and F1 scores [\citealt{zhang2023enhancing}]. RAG addresses the issue of factual accuracy in LLM outputs by retrieving contextually relevant external information. This technique enhances the model’s responses, especially in complex queries related to government financial data, ensuring both relevance and factual correctness in the generated content. While RAG shows promise in enhancing LLMs for public finance applications, challenges remain in ensuring the contextual relevance of retrieved data and maintaining performance consistency. Future developments in RAG and LLMs could further revolutionise data accessibility and decision-making processes in government finance.
Prompt engineering involves creating and refining prompts to improve interactions with LLMs. This technique involves adapting prompts to draw out the most precise, pertinent, or inventive responses from the models. The effectiveness of prompt engineering can significantly influence a model’s performance, proving essential for uses that include text generation and problem-solving. A thorough grasp of how the model processes and reacts to various inputs is necessary to enable developers to better utilise the model’s abilities. 
Utilising prompting and generative AI in financial decision-making provides numerous valuable applications that can improve efficiency, accuracy and strategic insights within the finance sector. Prompt engineering is crucial in transforming financial decision-making. The significance of its role in transforming the future of finance is highlighted by its effects on efficiency, risk management, customer satisfaction, and investment returns. This presentation of results using numerical data, graphical illustrations, and detailed analysis highlights the significant influence of prompt engineering on different financial aspects, offering a thorough insight into its effects on financial decision-making [\citealt{dhar2023analysis}].
Fine-tuning is the process of adapting a pre-trained model, typically by adding an extra layer, like a linear layer or multilayer perceptron (MLP), to optimise and tailor it for specific tasks, using both the broad knowledge gained during pre-training and a small number of task-specific parameters [\citealt{su2019generalizing}][\citealt{radiya2020fine}]. These models can provide more precise and context-aware analyses by fine-tuning LLMs on government financial datasets. The fine-tuned model is essential for budget analysis, fiscal policy evaluation, and financial forecasting, wherein a detailed understanding and high accuracy are critical.

\section{Experiments}
\subsection{Dataset Description}
This study gathered a comprehensive dataset from multiple reliable sources to create a solid basis for detailed LLM development. Open to the public, the data was procured from several vital organisations, each providing unique and essential information for the research. The Ministry of Finance provided various financial data and regulations [\citealt{ministry-of-finance-of-republic-of-indonesia-2023}]. Additional data was provided by Statistics Indonesia, which offers important statistical data and reports about Indonesia’s account balances [\citealt{statistics-indonesia-2023}].The IMF further enriched the dataset by supplying data on various economic indicators [\citealt{international-monetary-fund-2023}]. The data was collected in formats like PDF, CSV and TXT from 2003 to 2023. 
In addition to the data that was publicly accessible, a formal request was made to the Ministry of Finance to gain access to proprietary government datasets, specifically to retrieve data on financial transactions covering the period from 2014 to 2023. This aggregation presented a summary of the monthly financial transactions across various ministries, and the dataset included about 180,000 recorded transactions over the stated period. 
As part of the evaluation process, a comprehensive survey was conducted involving officials and experts from the Ministry of Finance of the Republic of Indonesia. Participants in the survey were asked not only to identify these shortcomings but also to provide the correct answers along with the sources of these answers. From this initiative, a total of 100 question and answer (Q\&A) entries were compiled.
For the fine-tuning of KemenkeuGPT, it was essential to gather an extensive collection of Q\&A pairs specifically dealing with Indonesian financial data. To achieve this, we closely examined the Q\&A section of the Ministry of Finance’s website to extract relevant data. Using a data scraping technique, we initially collected about 2,000 Q\&A pairs. This initial dataset underwent a detailed preprocessing phase to ensure the data was accurate and relevant for the model’s training purposes. Entries that did not meet the necessary standards for clarity, relevance, or completeness were carefully removed. As a result, the dataset was reduced to 1,688 valid Q\&A entries. Additional 1,299 Q\&A pairs were collected from The Directorate General of Customs and Excise.
As part of the continuous improvement process, KemenkeuGPT actively collects user feedback. Users can provide ratings and comments on the responses generated by KemenkeuGPT. This valuable feedback is gathered by the LangSmith platform and used to fine-tune the KemenkeuGPT model.

\subsection{LangChain Implementation}
This research implemented LangChain, a recent open-source software library that has gained attention in the AI community for offering solutions to streamline the development of custom AI applications using LLMs. The workings of LangChain are illustrated in Figure 2.

\begin{figure*}
	\centering 
	\includegraphics[width=0.75\textwidth]{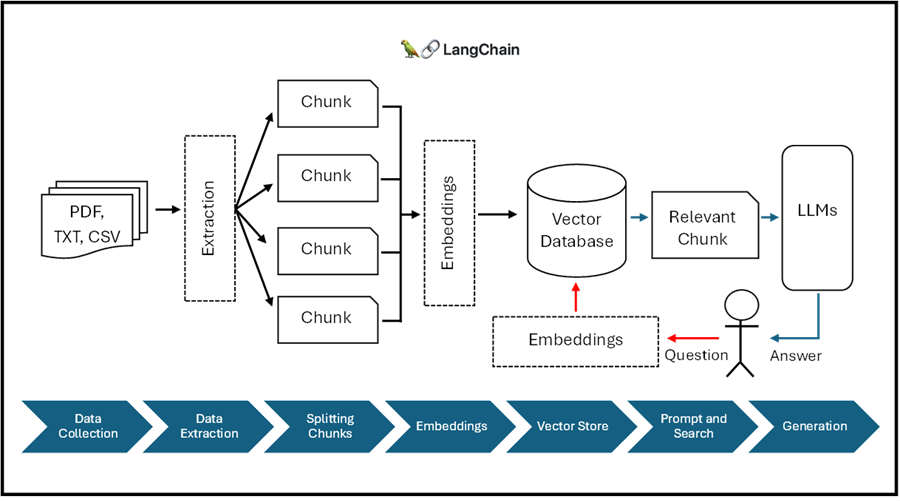}	
	\caption{Retrieval Augmented Generation (RAG) with LangChain.}
        \label{fig:img2}
\end{figure*}

Further exploration focused on using LangChain components, such as document loaders, chains, and text splitters, to create the most suitable environment for data extraction from the dataset. Since the dataset consisted of multiple document formats, another experiment was conducted using the DirectoryLoader. The DirectoryLoader is a component designed to handle the loading of documents from a specified directory efficiently, thus simplifying the management and processing of multiple files stored in one location.
Therefore, this study explored cloud solutions for document loading. Instead of storing the dataset in local storage, it was stored in Google Cloud Storage (GCS), and the GCSDirectoryLoader was utilised. The GCSDirectoryLoader, designed to interact with GCS, facilitates the loading and processing of files stored in a GCS bucket, making it a crucial tool for applications that require significant performance and scalability benefits. The complete comparison between the document’s loader is shown in Table 1.

\begin{table}[htbp]
\centering
\renewcommand{\arraystretch}{2}
\caption{Comparison of document loaders}
\begin{tabular}{|l|l|l|l|}
\hline
\textbf{Loader} & \textbf{Document Format} & \textbf{Files} & \textbf{Stored}  \\ \hline
PyPDFLoader & PDF & Single & Local \\ \hline 
PyPDFDirectoryLoader & PDF & Multiple & Local \\ \hline 
CSVLoader & CSV & Single & Local \\ \hline 
DirectoryLoader & PDF, CSV, etc & Multiple & Local \\ \hline 
GCSDirectoryLoader & PDF, CSV, etc & Multiple & Cloud \\ \hline 
\end{tabular}
\end{table}

During this phase of the project, we experimented chain types, including Stuff, MapReduce and Refine. Additionally, two distinct vector databases, namely ChromaDB and the Pinecone Vector Database, were compared to determine the most effective system for managing the financial data and regulations of the Indonesian government. The comparison of those vector databases is shown in Table 2.

\begin{table}[htbp]
\centering
\renewcommand{\arraystretch}{2}
\caption{Comparison of vector databases}
\begin{tabular}{|l|l|l|}
\hline
\textbf{Vector Database} & \textbf{Scalability} & \textbf{Cost} \\ \hline
Chroma DB & Depend on resources & Free (Open source) \\ \hline 
Pinecone & Highly Scalable & \$70.08 per month \\ \hline 
\end{tabular}
\end{table}

Efficient storage mechanisms are essential for handling large datasets, particularly when incorporating LLMs like LangChain for data analysis. In this project segment, we conducted experiments using Local Storage, Azure Blob Storage and Google Cloud Storage. The comparison of those storages is shown in Table 3.

\begin{table}[htbp]
\centering
\renewcommand{\arraystretch}{2}
\caption{Comparison of storages}
\begin{tabular}{|l|l|l|}
\hline
\textbf{Storage} & \textbf{Scalability} & \textbf{Cost} \\ \hline
Local Storage & Depend on resources & Depend on resources \\ \hline 
Azure Blob & Highly Scalable & \$0.023/GB/per month \\ \hline 
Google Cloud & Highly Scalable & \$0.020/GB/per month \\ \hline 
\end{tabular}
\end{table}

An intuitive and efficient user interface (UI) facilitates effective interaction with LLMs. In this project phase, LLM tools were developed using Streamlit, focusing on pivotal features, such as data visualisation, user interaction, and system responsiveness. After creating a user interface with Streamlit, which is accessible only via a web browser, a multi-platform application was developed with Flutter to enrich the user experience, enabling access to KemenkeuGPT across various platforms.

\subsection{Iterative Development}
The development and evaluation process for KemenkeuGPT comprised six phases:
\begin{itemize}
    \item Improvement through Experiments with LLMs.
    \item Improvement through Stakeholder Feedback.
    \item Improvement through Additional Data for RAG.
    \item Improvement through Prompt Engineering.
    \item Improvement through Fine-Tuning.
    \item Continuous Improvement through Human Feedback.
\end{itemize}

In the first phase, we implemented LangChain with RAG using seven different LLMs: OpenAI GPT-3.5 Turbo, Meta Llama 2 (Llama-2-7b-chat-hf), Amazon Bedrock Titan Text G1 - Express, Amazon Bedrock Titan Text G1 - Lite, Mistral 7B Instruct, Mixtral 8X7B Instruct, and AI21 Labs Jurassic-2 Ultra. Since OpenAI GPT-3.5 Turbo exhibited the best performance, we selected it as the base model for KemenkeuGPT. The performance of each model is detailed in Table IV.

\begin{table}[htbp]
\centering
\renewcommand{\arraystretch}{2}
\caption{The performance of 14 configured base LLMs}
\begin{tabular}{|l|l|l|l|l|l|}
\hline
\textbf{Base Model} & \textbf{Vector Darabase} & \textbf{Embedding}  & \textbf{Response (s)} & \textbf{Accuracy}\\ \hline
GPT-3.5 Turbo & ChromaDB & OpenAI & 1.43734 & 31\% \\ \hline
GPT-3.5 Turbo & Pinecone & OpenAI & 1.52909 & 35\% \\ \hline
Llama-2-7b-chat-hf & ChromaDB & OpenAI & 7.55333 & 28\% \\ \hline
Llama-2-7b-chat-hf & Pinecone & OpenAI & 6.65956 & 29\% \\ \hline
Titan Text G1 Express & ChromaDB & OpenAI & 3.16782 & 19\% \\ \hline
Titan Text G1 Express & Pinecone & OpenAI & 3.58585 & 22\% \\ \hline
Titan Text G1 Lite & ChromaDB & OpenAI & 2.85389 & 4\% \\ \hline
Titan Text G1 Lite & Pinecone & OpenAI & 3.31146 & 6\% \\ \hline
Mistral 7B & ChromaDB & OpenAI & 2.86586 & 22\% \\ \hline
Mistral 7B & Pinecone & OpenAI & 3.07261 & 27\% \\ \hline
Mixtral 8X7B & ChromaDB & OpenAI & 2.01232 & 27\% \\ \hline
Mixtral 8X7B & Pinecone & OpenAI & 2.02417 & 30\% \\ \hline
Jurassic-2 Ultra & ChromaDB & OpenAI & 1.53169 & 12\% \\ \hline
Jurassic-2 Ultra & Pinecone & OpenAI & 1.74307 & 13\% \\ \hline
\end{tabular}
\end{table}

In the second phase, we conducted a survey to gather feedback. The survey, conducted from March 21st to April 6th, 2024, invited staff members of the Ministry of Finance of the Republic of Indonesia to participate. Thematic analysis, following the framework proposed by Braun and Clarke [\citealt{braun2006using}], was utilised in this study. This survey identified five key themes essential for enhancing KemenkeuGPT’s performance: increasing its knowledge base, ensuring language consistency, improving model accuracy, providing sources for answers, and enhancing its UI.

In the third phase, in response to stakeholder feedback, more than 1,000 documents were added to the database designated for KemenkeuGPT’s use. These documents were systematically gathered from the Ministry of Finance’s website using a web scraping method. The performance of improved model is shown in Table V.

\begin{table}[htbp]
\centering
\renewcommand{\arraystretch}{2}
\caption{The performance of GPT-3.5 Turbo after adding more data}
\begin{tabular}{|l|l|l|l|l|l|}
\hline
\textbf{Model} & \textbf{Vector Darabase} & \textbf{Embedding}  & \textbf{Response (s)} & \textbf{Accuracy}\\ \hline
GPT-3.5 Turbo & Pinecone & OpenAI & 2.26909 & 42\% \\ \hline
\end{tabular}
\end{table}

In the fourth phase, we implemented prompt engineering by creating a LangChain agent. This agent can transform the output format into a dictionary, which can then be extracted and transformed into a table or chart format using the Pandas library in Python. Additionally, by implementing prompt engineering techniques, the model was instructed to recognize the input language and generate a response in the same language. This adjustment ensured consistent and clear communication, regardless of the input language. The performance of enhanched model is shown in Table VI.

\begin{table}[htbp]
\centering
\renewcommand{\arraystretch}{2}
\caption{The performance of GPT-3.5 Turbo after prompt engineering.}
\begin{tabular}{|l|l|l|l|l|l|}
\hline
\textbf{Model} & \textbf{Vector Darabase} & \textbf{Embedding}  & \textbf{Response (s)} & \textbf{Accuracy}\\ \hline
GPT-3.5 Turbo & Pinecone & OpenAI & 10.92692 & 60\% \\ \hline
\end{tabular}
\end{table}

In the fifth phase, we fine-tuned the base model to enhance KemenkeuGPT’s performance. Based on previous experiments, it was determined that OpenAI’s GPT-3.5 Turbo significantly outperformed the other six models that were tested. Consequently, it was selected as the base model and fine-tuned using the collected Q\&A data. The performance of fine tuned model is shown in Table VII.

\begin{table}[htbp]
\centering
\renewcommand{\arraystretch}{2}
\caption{The performance of KemenkeuGPT Model after fine tuning.}
\begin{tabular}{|l|l|l|l|l|l|}
\hline
\textbf{Model} & \textbf{Vector Darabase} & \textbf{Embedding}  & \textbf{Response (s)} & \textbf{Accuracy}\\ \hline
ft:gpt-3.5-turbo-0125:personal::9RAg3mNo & Pinecone & OpenAI & 10.52602 & 61\% \\ \hline
\end{tabular}
\end{table}

In the sixth phase, we implemented a feedback feature that allowed users to assess KemenkeuGPT’s responses as shown in Figure 3. Each piece of feedback was collected and evaluated by experts and used either for the fine-tuning process or for the integration of new data into the RAG data source.

\begin{figure}
	\centering 
	\includegraphics{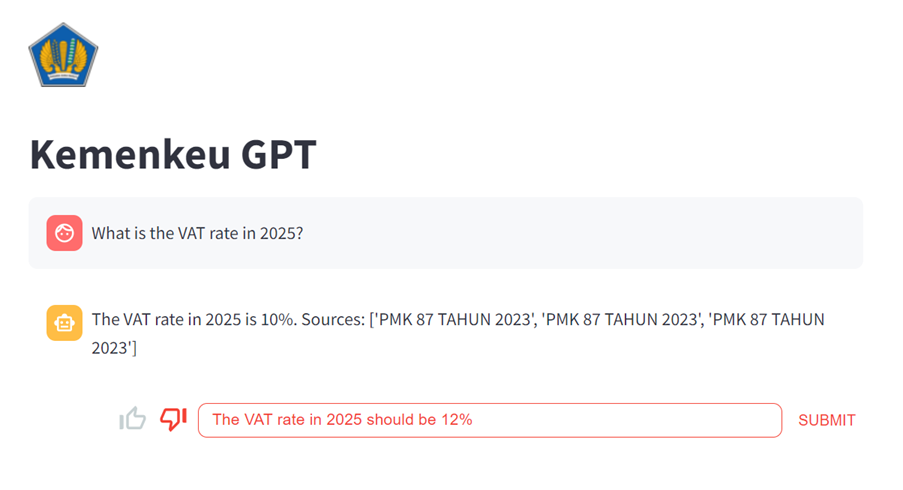}	
	\caption{KemenkeuGPT feedback feature for continues improvement.}
        \label{fig:img3}
\end{figure}

\subsection{Evaluation}
This study employs a human evaluation method, engaging collaborators from various departments within the Ministry of Finance of the Republic of Indonesia. These collaborators, possessing diverse domain expertise in public finance, play a pivotal role in assessing the responses generated by the LLM. Their task is to closely assess the accuracy of the LLM’s outputs, specifically in the context of public finance decision-making. A survey will be conducted to gather feedback from staff and officials in the Ministry of Finance. Alongside the manual evaluation conducted by human experts, the study also implements an LLM-based evaluation using LangChain string evaluator. This method involves a systematic and detailed assessment of the responses generated by the LLMs, ensuring a comprehensive examination of their correctness and reliability. Furthermore, the study includes a comparative analysis between KemenkeuGPT and other LLMs utilising the RAGAS framework. The RAGAS framework evaluates faithfulness, correctness, precision, and recall. Faithfulness measures how factually consistent the generated answer is with the given context, calculated from the answer and the retrieved context, with scores ranging from 0 to 1. Higher scores indicate better faithfulness, as shown in equation 1. Correctness assesses the accuracy of the generated answer as compared to the ground truth. This metric also ranges from 0 to 1, with higher scores reflecting a closer match between the generated answer and the ground truth, as detailed in equation 2. Context precision measures the extent to which all relevant items from the ground truth are ranked higher among the retrieved contexts. This metric, determined using the question, ground truth, and contexts, produces scores between 0 and 1, with higher scores indicating greater precision, as demonstrated in equation 3. On the other hand, context recall assesses how well the retrieved context aligns with the annotated answer which is regarded as the ground truth. This metric is calculated by comparing the ground truth with the retrieved context, also yielding scores between 0 and 1, with higher scores representing better performance, as illustrated in equation 4.

\begin{equation}
\text{Faithfulness} = \frac{|\text{Claims answer supported by context}|}{|\text{Total number of claims in answer}|}
\label{eq:faithfulness_score}
\end{equation}

\begin{equation}
\text{Correctness} = \frac{|\text{TP}|}{(|\text{TP}| + 0.5 \times (|\text{FP}| + |\text{FN}|))}
\label{eq:f1_score}
\end{equation}

\begin{equation}
\text{Precision@K} = \frac{\sum_{k=1}^{K} (\text{Precision@k} \times v_k)}{\text{Number of items in the top } K \text{ results}}
\label{eq:context_precision_k}
\end{equation}

\begin{equation}
\text{Recall} = \frac{|\text{GT sentences attributed to context}|}{|\text{Number of sentences in GT}|}
\label{eq:context_recall}
\end{equation}

\section{Results and Discussions}
After conducting several experiments and configurations, the results from each phase were systematically recorded and analysed. According to the human evaluation, the accuracy of KemenkeuGPT increased with each improvement. Since KemenkeuGPT is a continuous improvement model that uses human feedback, its performance is expected to keep improving in the future. By the end of this research, the accuracy of KemenkeuGPT reached 61\%. This result was achieved after iterative development using the base model, adding more data for RAG, prompt engineering and fine-tuning methods. The outcomes of each stage are hereby presented in Table VIII and Figure 4. Initially, the accuracy was only 35\%. After adding more data for RAG, it increased to 42\%. Exploring prompt engineering raised it to 60\% and fine-tuning brought it to 61\%. Even though the result is still below 70\%, KemenkeuGPT has shown significant improvement, which is a positive sign for future enhancements.

\begin{table}[htbp]
\centering
\renewcommand{\arraystretch}{2}
\caption{Accuracy of KemenkeuGPT after iterative improvement}
\begin{tabular}{|l|l|l|l|l|}
\hline
\textbf{Performance} & \textbf{Base Model} & \textbf{RAG} & \textbf{Prompt Engineering}  & \textbf{Fine Tuning} \\ \hline
Accuracy & 35\% & 42\% & 60\% & 61\% \\ \hline 
\end{tabular}
\end{table}

\begin{figure}
	\centering 
	\includegraphics{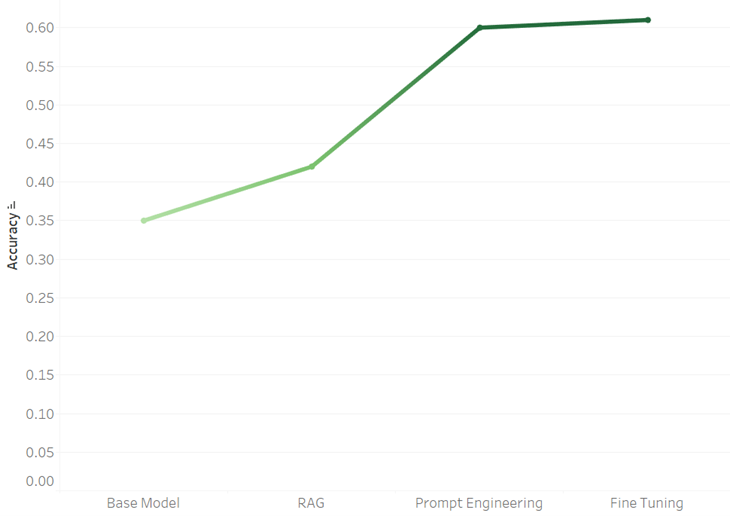}	
	\caption{Accuracy of KemenkeuGPT after iterative improvement.}
        \label{fig:img4}
\end{figure}

The results indicate that response time increases as more data are added. However, fine-tuning the model slightly decreases it. The results are shown in Table IX and Figure 5. As more data were added and the algorithm and prompts were refined, response time increased significantly. However, after fine-tuning, there was a slight decrease of a few milliseconds, suggesting that fine-tuning can reduce the response time.

\begin{table}[htbp]
\centering
\renewcommand{\arraystretch}{2}
\caption{Response of KemenkeuGPT after iterative improvement}
\begin{tabular}{|l|l|l|l|l|}
\hline
\textbf{Performance} & \textbf{Base Model} & \textbf{RAG} & \textbf{Prompt Engineering}  & \textbf{Fine Tuning} \\ \hline
Response Time (s) & 1.52909 & 2.26909 & 10.92692 & 10.52602 \\ \hline
\end{tabular}
\end{table}

\begin{figure}
	\centering 
	\includegraphics{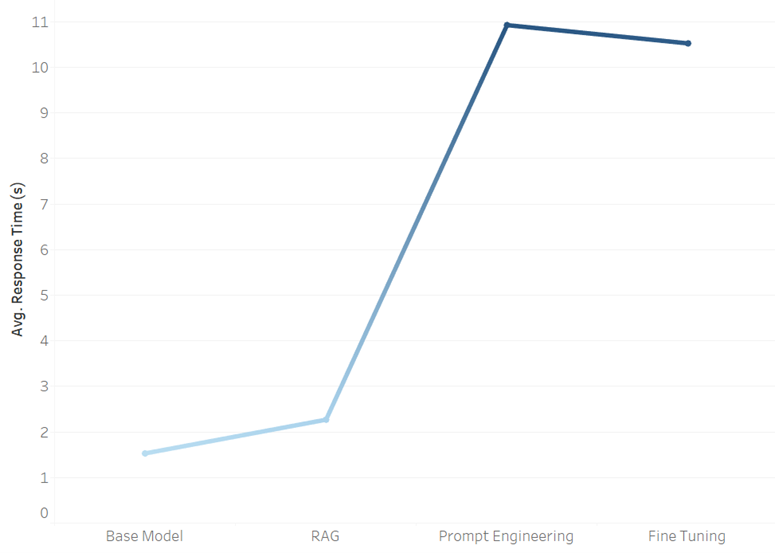}	
	\caption{Response of KemenkeuGPT after iterative improvement.}
        \label{fig:img5}
\end{figure}

Based on the LLM-based evaluation, KemenkeuGPT achieved 64\% correctness, while its base model had 48\% correctness,  as shown in Figure 6. Although KemenkeuGPT’s correctness was not very high, the iterative process improved its correctness by 16\%, which indicated a notable improvement.

\begin{figure}
	\centering 
	\includegraphics[width=0.8\textwidth]{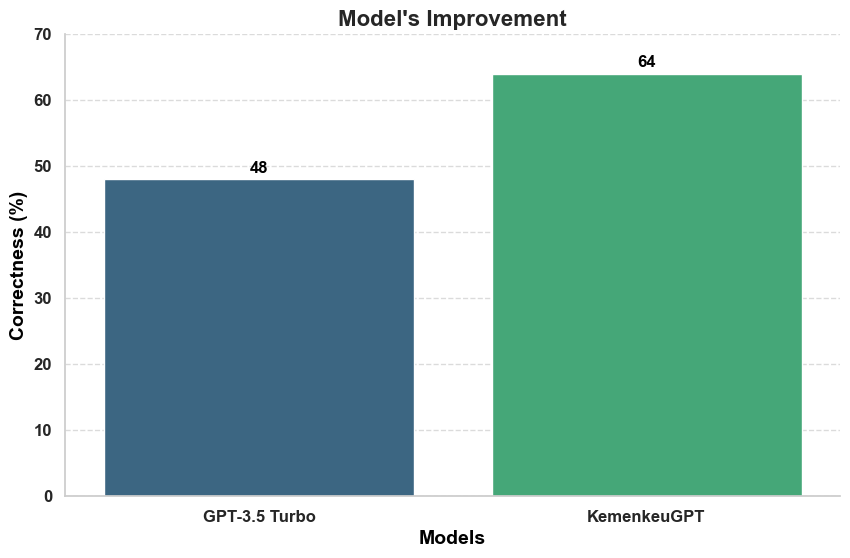}	
	\caption{Performance of the base model and KemenkeuGPT after improvement.}
        \label{fig:img6}
\end{figure}

Finally, benchmarking with the RAGAS framework showed that KemenkeuGPT slightly outperformed the other seven LLMs, as shown in Table X and Figure 7. KemenkeuGPT achieved 44\% correctness, 73\% faithfulness, 40\% precision and 60\% recall. Although the results were still low, all scores were higher than those of the other LLMs. This indicates that KemenkeuGPT performs better in answering questions related to Indonesian financial data and regulations.

\begin{table}[htbp]
\centering
\renewcommand{\arraystretch}{2}
\caption{Performance matrix of KemenkeuGPT and other seven models}
\begin{tabular}{|l|l|l|l|l|}
\hline
\textbf{Model} & \textbf{Correctness} & \textbf{Faithfullness} & \textbf{Precision}  & \textbf{Recall} \\ \hline
KemenkeuGPT & 0.44 & 0.73 & 0.40 & 0.60 \\ \hline
OpenAI GPT-3.5 Turbo & 0.42 & 0.47 & 0.20 & 0.40 \\ \hline
Llama-2-7b-chat-hf & 0.39 & 0.36 & 0.30 & 0.40 \\ \hline
Titan Text G1 - Express & 0.27 & 0.46 & 0.30 & 0.40 \\ \hline
Titan Text G1 - Lite & 0.19 & 0.60 & 0.30 & 0.40 \\ \hline
Mistral 7B Instruct & 0.38 & 0.66 & 0.30 & 0.40 \\ \hline
Mixtral 8X7B Instruct & 0.32 & 0.71 & 0.20 & 0.40 \\ \hline
Jurassic-2 Ultra & 0.25 & 0.20 & 0.30 & 0.40 \\ \hline
\end{tabular}
\end{table}

\begin{figure*}
	\centering 
	\includegraphics[width=0.85\textwidth]{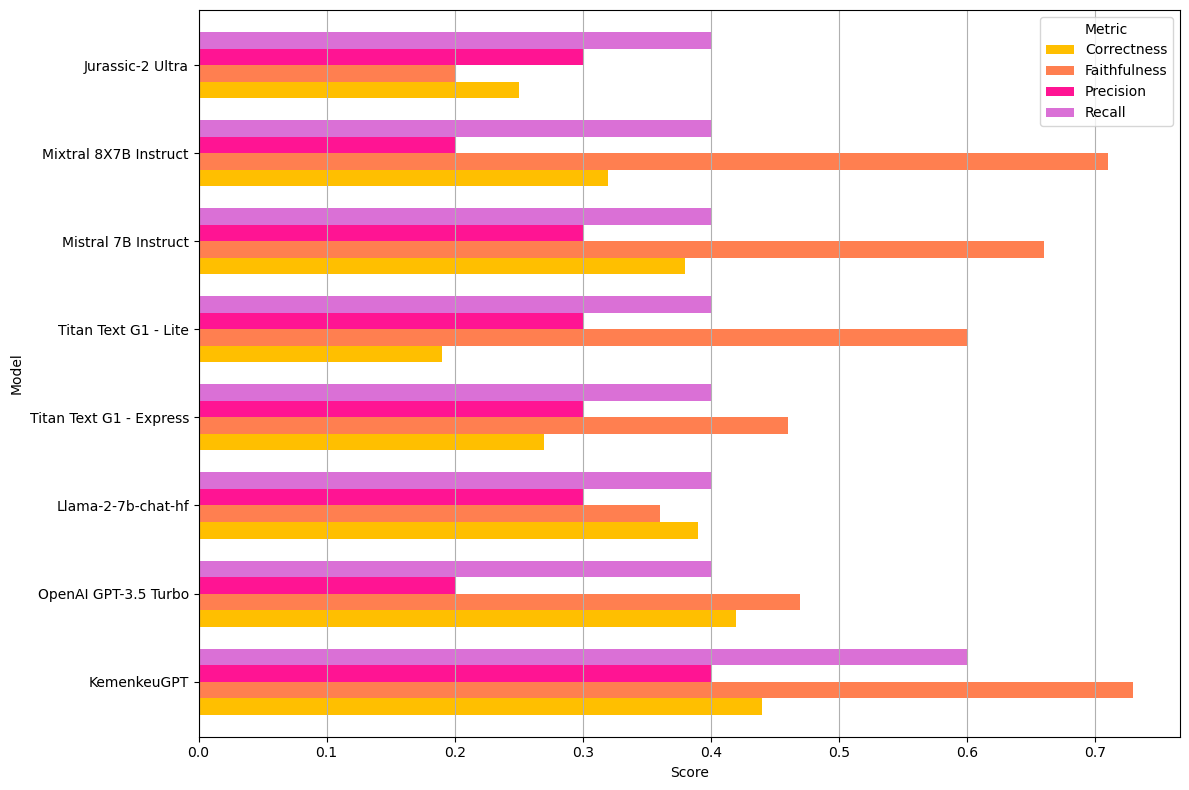}	
	\caption{Performance matrix of KemenkeuGPT and seven other models}
        \label{fig:img7}
\end{figure*}

Additionally, an interview with an expert from the Ministry of Finance indicated that KemenkeuGPT has the potential to become an essential tool for decision-making. It can extract information and insights from Indonesian financial data and regulations, which would benefit staff, especially in customer service, in quickly obtaining accurate information. Moreover, KemenkeuGPT is advantageous for the Board of Directors in The Ministry of Finance to obtain information specific to Indonesian financial data and regulations, unlike other LLMs like ChatGPT, which only provide general answers.

This suggests that KemenkeuGPT could become a promising tool for customer service, serving as the first point of contact and shifting the customer service paradigm from human agents to AI-powered solutions.

\section{Conclusions}
Large Language Models (LLMs) have been studied for its application in extracting information and insights from Indonesian financial data and regulations using the LangChain framework. The process is iterative, involving Retrieval-Augmented Generation (RAG), prompt engineering and fine-tuning. The experimental results show that this iterative process improves KemenkeuGPT's performance. Although KemenkeuGPT's performance is not exceptionally high, it still outperforms its base model and six other models in answering Indonesian financial data and regulations questions. Additionally, since KemenkeuGPT is a continuous improvement model with human feedback, its performance is expected to improve over time. An expert interview highlighted KemenkeuGPT's potential to become an essential tool for decision-making. These results indicate that KemenkeuGPT, as an LLM, can enhance the decision-making process by extracting information and insights from Indonesian financial data and regulations.

However, despite KemenkeuGPT's superior performance compared to other models, there are limitations. First, KemenkeuGPT currently has access to only partial data on Indonesian financial data and regulations. One reason is that the Ministry of Finance rejected our requests for additional data. Furthermore, in compliance with the terms of a non-disclosure agreement, our analysis is confined to an aggregate overview of government financial transactions. This restriction prevents detailed analysis of individual transactions, which could offer more profound insights. Consequently, using partial and aggregate data might limit the study's scope and the depth of its findings. Secondly, KemenkeuGPT cannot update its dataset, so new data must be manually uploaded to the database. Thirdly, continuous human feedback evaluation is required due to the rigidity of financial data and regulations, ensuring the content is accurate and not misleading. This process needs to be manually assessed by the Ministry of Finance experts, which can be time-consuming. Additionally, the current user interface of KemenkeuGPT is limited to a web view from Flutter and could be improved.

Several enhancements could refine the research for future work. Expanding the dataset with additional resources would improve the RAG process and model performance. Including detailed transaction data would provide users with richer information for better decision-making. Integrating with the Ministry of Finance data centre can enable real-time data updates, eliminating the need for manual refreshes. Additionally, exploring alternative base models and diversifying experimental approaches, such as using different combinations of base models, embeddings and vector databases or experimenting with enhanced prompts, can improve LLM performance.

\section*{Acknowledgment}

We extend our gratitude to the Indonesia Endowment Fund for Education (LPDP) of the Ministry of Finance of the Republic of Indonesia for funding this research. Additionally, we sincerely thank Saiful Islam, Faried Zamachsari, and Decci Farroza for their evaluation and insightful feedback on KemenkeuGPT.

\bibliographystyle{unsrtnat}



\begin{thebibliography}{30}
\providecommand{\natexlab}[1]{#1}
\providecommand{\url}[1]{\texttt{#1}}
\expandafter\ifx\csname urlstyle\endcsname\relax
  \providecommand{\doi}[1]{doi: #1}\else
  \providecommand{\doi}{doi: \begingroup \urlstyle{rm}\Url}\fi

\bibitem[van Ooijen et~al.(2019)van Ooijen, Ubaldi, and Welby]{van2019data}
Charlotte van Ooijen, Barbara Ubaldi, and Benjamin Welby.
\newblock A data-driven public sector: Enabling the strategic use of data for productive, inclusive and trustworthy governance.
\newblock \emph{OECD Working Papers on Public Governance}, 2019.

\bibitem[Canrakerta et~al.(2021)Canrakerta, Hutabarat, Rahadian, Sirait, and Wicaksana]{canrakerta2021membangun}
Canrakerta Canrakerta, Dody~Dharma Hutabarat, Dimas Rahadian, Lysa~Novita Sirait, and Lazuardi~Zulfikar Wicaksana.
\newblock \emph{Building a Data Culture in the Ministry of Finance}.
\newblock Central Transformation Office, Sekretariat Jenderal, Kementerian Keuangan, 2021.

\bibitem[Chen et~al.(2024)Chen, Gasc{\'o}-Hernandez, and Esteve]{chen2024adoption}
Tzuhao Chen, Mila Gasc{\'o}-Hernandez, and Marc Esteve.
\newblock The adoption and implementation of artificial intelligence chatbots in public organizations: Evidence from us state governments.
\newblock \emph{The American Review of Public Administration}, 54\penalty0 (3):\penalty0 255--270, 2024.

\bibitem[Porreca et~al.(2018)Porreca, Leotta, Mecella, Vassos, and Catarci]{porreca2018accessing}
Simone Porreca, Francesco Leotta, Massimo Mecella, Stavros Vassos, and Tiziana Catarci.
\newblock Accessing government open data through chatbots.
\newblock In \emph{Current Trends in Web Engineering: ICWE 2017 International Workshops, Liquid Multi-Device Software and EnWoT, practi-O-web, NLPIT, SoWeMine, Rome, Italy, June 5-8, 2017, Revised Selected Papers 17}, pages 156--165. Springer, 2018.

\bibitem[Provost and Fawcett(2013)]{provost2013data}
Foster Provost and Tom Fawcett.
\newblock Data science and its relationship to big data and data-driven decision making.
\newblock \emph{Big data}, 1\penalty0 (1):\penalty0 51--59, 2013.

\bibitem[Sarker(2021)]{sarker2021data}
Iqbal~H Sarker.
\newblock Data science and analytics: an overview from data-driven smart computing, decision-making and applications perspective.
\newblock \emph{SN Computer Science}, 2\penalty0 (5):\penalty0 377, 2021.

\bibitem[Ojokoh et~al.(2020)Ojokoh, Samuel, Omisore, Sarumi, Idowu, Chimusa, Darwish, Adekoya, and Katsriku]{ojokoh2020big}
Bolanle~A Ojokoh, Oluwarotimi~W Samuel, Olatunji~M Omisore, Oluwafemi~A Sarumi, Peter~A Idowu, Emile~R Chimusa, Ashraf Darwish, Adebayo~F Adekoya, and Ferdinand~A Katsriku.
\newblock Big data, analytics and artificial intelligence for sustainability, 2020.

\bibitem[Charles et~al.(2022)Charles, Rana, and Carter]{charles2022artificial}
Vincent Charles, Nripendra~P Rana, and Lemuria Carter.
\newblock Artificial intelligence for data-driven decision-making and governance in public affairs, 2022.

\bibitem[Zhao et~al.(2023)Zhao, Zhou, Li, Tang, Wang, Hou, Min, Zhang, Zhang, Dong, et~al.]{zhao2023survey}
Wayne~Xin Zhao, Kun Zhou, Junyi Li, Tianyi Tang, Xiaolei Wang, Yupeng Hou, Yingqian Min, Beichen Zhang, Junjie Zhang, Zican Dong, et~al.
\newblock A survey of large language models.
\newblock \emph{arXiv preprint arXiv:2303.18223}, 2023.

\bibitem[OpenAI(2023)]{openai2023gpt}
R~OpenAI.
\newblock Gpt-4 technical report. arxiv 2303.08774.
\newblock \emph{View in Article}, 2\penalty0 (5), 2023.

\bibitem[Touvron et~al.(2023)Touvron, Lavril, Izacard, Martinet, Lachaux, Lacroix, Rozi{\`e}re, Goyal, Hambro, Azhar, et~al.]{touvron2023llama}
Hugo Touvron, Thibaut Lavril, Gautier Izacard, Xavier Martinet, Marie-Anne Lachaux, Timoth{\'e}e Lacroix, Baptiste Rozi{\`e}re, Naman Goyal, Eric Hambro, Faisal Azhar, et~al.
\newblock Llama: Open and efficient foundation language models.
\newblock \emph{arXiv preprint arXiv:2302.13971}, 2023.

\bibitem[Chowdhery et~al.(2023)Chowdhery, Narang, Devlin, Bosma, Mishra, Roberts, Barham, Chung, Sutton, Gehrmann, et~al.]{chowdhery2023palm}
Aakanksha Chowdhery, Sharan Narang, Jacob Devlin, Maarten Bosma, Gaurav Mishra, Adam Roberts, Paul Barham, Hyung~Won Chung, Charles Sutton, Sebastian Gehrmann, et~al.
\newblock Palm: Scaling language modeling with pathways.
\newblock \emph{Journal of Machine Learning Research}, 24\penalty0 (240):\penalty0 1--113, 2023.

\bibitem[Thoppilan et~al.(2022)Thoppilan, De~Freitas, Hall, Shazeer, Kulshreshtha, Cheng, Jin, Bos, Baker, Du, et~al.]{thoppilan2022lamda}
Romal Thoppilan, Daniel De~Freitas, Jamie Hall, Noam Shazeer, Apoorv Kulshreshtha, Heng-Tze Cheng, Alicia Jin, Taylor Bos, Leslie Baker, Yu~Du, et~al.
\newblock Lamda: Language models for dialog applications.
\newblock \emph{arXiv preprint arXiv:2201.08239}, 2022.

\bibitem[Taylor et~al.(2022)Taylor, Kardas, Cucurull, Scialom, Hartshorn, Saravia, Poulton, Kerkez, and Stojnic]{taylor2022galactica}
Ross Taylor, Marcin Kardas, Guillem Cucurull, Thomas Scialom, Anthony Hartshorn, Elvis Saravia, Andrew Poulton, Viktor Kerkez, and Robert Stojnic.
\newblock Galactica: A large language model for science.
\newblock \emph{arXiv preprint arXiv:2211.09085}, 2022.

\bibitem[Devlin et~al.(2018)Devlin, Chang, Lee, and Toutanova]{devlin2018bert}
Jacob Devlin, Ming-Wei Chang, Kenton Lee, and Kristina Toutanova.
\newblock Bert: Pre-training of deep bidirectional transformers for language understanding.
\newblock \emph{arXiv preprint arXiv:1810.04805}, 2018.

\bibitem[Chakrabarty et~al.(2019)Chakrabarty, Hidey, and McKeown]{chakrabarty2019imho}
Tuhin Chakrabarty, Christopher Hidey, and Kathleen McKeown.
\newblock Imho fine-tuning improves claim detection.
\newblock \emph{arXiv preprint arXiv:1905.07000}, 2019.

\bibitem[Liu et~al.(2023)Liu, Wang, and Zha]{liu2023fingpt}
Xiao-Yang Liu, Guoxuan Wang, and Daochen Zha.
\newblock Fingpt: Democratizing internet-scale data for financial large language models.
\newblock \emph{arXiv preprint arXiv:2307.10485}, 2023.

\bibitem[Wu et~al.(2023)Wu, Irsoy, Lu, Dabravolski, Dredze, Gehrmann, Kambadur, Rosenberg, and Mann]{wu2023bloomberggpt}
Shijie Wu, Ozan Irsoy, Steven Lu, Vadim Dabravolski, Mark Dredze, Sebastian Gehrmann, Prabhanjan Kambadur, David Rosenberg, and Gideon Mann.
\newblock Bloomberggpt: A large language model for finance.
\newblock \emph{arXiv preprint arXiv:2303.17564}, 2023.

\bibitem[Huang et~al.(2023)Huang, Wang, and Yang]{huang2023finbert}
Allen~H Huang, Hui Wang, and Yi~Yang.
\newblock Finbert: A large language model for extracting information from financial text.
\newblock \emph{Contemporary Accounting Research}, 40\penalty0 (2):\penalty0 806--841, 2023.

\bibitem[Topsakal and Akinci(2023)]{topsakal2023creating}
Oguzhan Topsakal and Tahir~Cetin Akinci.
\newblock Creating large language model applications utilizing langchain: A primer on developing llm apps fast.
\newblock In \emph{International Conference on Applied Engineering and Natural Sciences}, volume~1, pages 1050--1056, 2023.

\bibitem[Cai et~al.(2022)Cai, Wang, Liu, and Shi]{cai2022recent}
Deng Cai, Yan Wang, Lemao Liu, and Shuming Shi.
\newblock Recent advances in retrieval-augmented text generation.
\newblock In \emph{Proceedings of the 45th international ACM SIGIR conference on research and development in information retrieval}, pages 3417--3419, 2022.

\bibitem[Lewis et~al.(2020)Lewis, Perez, Piktus, Petroni, Karpukhin, Goyal, K{\"u}ttler, Lewis, Yih, Rockt{\"a}schel, et~al.]{lewis2020retrieval}
Patrick Lewis, Ethan Perez, Aleksandra Piktus, Fabio Petroni, Vladimir Karpukhin, Naman Goyal, Heinrich K{\"u}ttler, Mike Lewis, Wen-tau Yih, Tim Rockt{\"a}schel, et~al.
\newblock Retrieval-augmented generation for knowledge-intensive nlp tasks.
\newblock \emph{Advances in Neural Information Processing Systems}, 33:\penalty0 9459--9474, 2020.

\bibitem[Zhang et~al.(2023)Zhang, Yang, Zhou, Ali~Babar, and Liu]{zhang2023enhancing}
Boyu Zhang, Hongyang Yang, Tianyu Zhou, Muhammad Ali~Babar, and Xiao-Yang Liu.
\newblock Enhancing financial sentiment analysis via retrieval augmented large language models.
\newblock In \emph{Proceedings of the Fourth ACM International Conference on AI in Finance}, pages 349--356, 2023.

\bibitem[Dhar et~al.(2023)Dhar, Datta, and Das]{dhar2023analysis}
Abhik Dhar, Aritra Datta, and Soma Das.
\newblock Analysis on enhancing financial decision-making through prompt engineering.
\newblock In \emph{2023 7th International Conference on Electronics, Materials Engineering \& Nano-Technology (IEMENTech)}, pages 1--5. IEEE, 2023.

\bibitem[Su et~al.(2019)Su, Xu, Winata, Xu, Kim, Liu, and Fung]{su2019generalizing}
Dan Su, Yan Xu, Genta~Indra Winata, Peng Xu, Hyeondey Kim, Zihan Liu, and Pascale Fung.
\newblock Generalizing question answering system with pre-trained language model fine-tuning.
\newblock In \emph{Proceedings of the 2nd workshop on machine reading for question answering}, pages 203--211, 2019.

\bibitem[Radiya-Dixit and Wang(2020)]{radiya2020fine}
Evani Radiya-Dixit and Xin Wang.
\newblock How fine can fine-tuning be? learning efficient language models.
\newblock In \emph{International Conference on Artificial Intelligence and Statistics}, pages 2435--2443. PMLR, 2020.

\bibitem[of~Finance of Republic~of Indonesia(2023)]{ministry-of-finance-of-republic-of-indonesia-2023}
Ministry of~Finance of Republic~of Indonesia.
\newblock {Ministry of Finance Regulation Information Network}, 2023.
\newblock URL \url{https://jdih.kemenkeu.go.id/en/home}.

\bibitem[Indonesia(2023)]{statistics-indonesia-2023}
Statistics Indonesia.
\newblock {Statistics Indonesia Publication}, 2023.
\newblock URL \url{https://www.bps.go.id/publication.html}.

\bibitem[Fund(2023)]{international-monetary-fund-2023}
International~Monetary Fund.
\newblock {IMF data}, 7 2023.
\newblock URL \url{https://www.imf.org/en/Data}.

\bibitem[Braun and Clarke(2006)]{braun2006using}
Virginia Braun and Victoria Clarke.
\newblock Using thematic analysis in psychology.
\newblock \emph{Qualitative research in psychology}, 3\penalty0 (2):\penalty0 77--101, 2006.

\end{thebibliography}





\end{document}